\documentclass[11pt,a4paper]{article}
\usepackage[hyperref]{eacl2021}
\usepackage{times}
\usepackage{latexsym}

\usepackage{times}
\usepackage{latexsym}
\usepackage{multirow}
\usepackage{graphicx}
\usepackage{framed}
\usepackage{subfig}
\usepackage{amsmath}
\usepackage{url}

\usepackage{microtype}

 \usepackage{todonotes}

\newcommand{\Note}[2]{} 
\newcommand{\SideNote}[2]{} 
\renewcommand{\Note}[2]{\todo[color=#1,size=\small, inline=true]{#2}} 
\renewcommand{\SideNote}[2]{\todo[color=#1,size=\small]{#2}} %

\aclfinalcopy 

\makeatletter
\newcommand{\printfnsymbol}[1]{%
  \textsuperscript{\@fnsymbol{#1}}%
}
\makeatother

\title{``Laughing at you or with you'': The Role of Sarcasm\\ in Shaping the Disagreement Space}

\author{Debanjan Ghosh\thanks{~~Equal Contribution.}  \textsuperscript{1}, 
  Ritvik Shrivastava\printfnsymbol{1}\textsuperscript{2, 3} \textbf{and}
  Smaranda Muresan\textsuperscript{3}
  \\ 
  \textsuperscript{1}Educational Testing Service \\
  \textsuperscript{2}MindMeld, Cisco Systems \\
  \textsuperscript{3}Data Science Institute, Columbia University\\
  {\tt dghosh@ets.org},
  {\tt ritvshri@cisco.com},
  {\tt \{rs3868, smara\}@columbia.edu}
  }

\date{}

\begin{document}
\maketitle
\begin{abstract}
Detecting arguments in online interactions is useful to understand how conflicts arise and get resolved. Users often use figurative language, such as sarcasm, either as persuasive devices or to attack the opponent by an ad hominem argument. To further our understanding of the role of sarcasm in shaping the disagreement space, we present a thorough experimental setup using a corpus annotated with both argumentative moves (agree/disagree) and sarcasm. We exploit joint modeling in terms of (a) applying discrete features that are useful in detecting sarcasm to the task of argumentative relation classification (agree/disagree/none), and (b) multitask learning for argumentative relation classification and sarcasm detection using deep learning architectures (e.g., dual Long Short-Term Memory (LSTM) with hierarchical attention and Transformer-based architectures). 
We demonstrate that modeling sarcasm improves the argumentative relation classification task (agree/disagree/none) in all setups.   

\end{abstract}

\section{Introduction} \label{section:introduction}

User-generated conversational data such as discussion forums provide a wealth of naturally occurring arguments. 
The ability to automatically detect and classify argumentative relations (e.g., agree/disagree)  in threaded discussions is useful to understand how collective opinions form, how conflict arises and is resolved \citep{eemeren1993reconstructing,abbott2011,walker2012corpus,misra2013,ghosh2014analyzing,rosenthal2015couldn,stede2018argumentation}. 
\begin{table}
\centering
\small
\begin{tabular}{ p{1.2cm}p{5.6cm} }
\hline
 Arg. Rel.  &  Turn Pairs \\
 \hline
   & {\bf Prior Turn:} Today, no informed creationist would deny natural selection.  \\
    $Agree$
   & {\bf Current Turn:} Seeing how this was proposed over a century and a half ago by Darwin, what took the creationists so long to catch up?\\
\hline
&  {\bf Prior Turn:} Personally I wouldn’t own a
gun for self defense because I am just not
that big of a sissy.\\
$Disagree$
& {\bf Current Turn:} Because taking responsibility for ones own safety is certainly a sissy
thing to do?\\
 \cline{2-2}
&  {\bf Prior Turn:}  I'm not surprised that no one on your side of the debate would correct you, but wolves and dogs are both members of the same species. The Canid species.  \\
  \multirow{4}{*} &   {\bf  Current Turn:} Wow, you 're even wrong when you get away from your precious Bible and try to sound scientific. \\
\hline 
& {\bf Prior Turn:} The hand of God kept me from serious harm. Maybe He has a plan for me. \\
$None$ & {\bf Current Turn:} You better hurry up . Are n't you like 113 years old. \\
\hline
\end{tabular}
\caption{Sarcastic turns that disagree, agree or have no argumentative relation with their prior turns.}
\label{table:sarcroleexample}
\vspace{-5mm}
\end{table}
Linguistic and argumentation theories have thoroughly studied the use of sarcasm in argumentation, including its effectiveness as a persuasive device or as a means to express an ad hominem fallacy (attacking the opponent instead of her/his argument) \cite{tindale1987use,vanEemren1992, gibbs2005irony,Averbeck2013}. 
We propose an experimental setup to further our understanding of the role of sarcasm in shaping up the disagreement space in online interactions.
The disagreement space, defined in the context of the dialogical perspective on argumentation, is seen as the speech acts initiating the difference of opinions that argumentation is intended to resolve \cite{Jackson1992,eemeren1993reconstructing}. Our study is based on the Internet Argument Corpus (IAC) introduced by \newcite{abbott2011} that contains online discussions annotated for the presence/absence and the type of an argumentative move (agree/disagree/none) as well as the presence/absence of sarcasm. 
Consider the dialogue turns from IAC in Table \ref{table:sarcroleexample},  where the \textbf{current turn} (henceforth, $ct$) is a sarcastic response to the \textbf{prior turn} (henceforth, $pt$). These dialogue moves can be argumentative (agree/disagree) or not argumentative (none). The argumentative move can express agreement (first example) or disagreement (the second example is an undercutter, while the third example is an ad hominem attack). The fourth example, although sarcastic, it is not argumentative. 
It can be noticed that none of the current turns contain explicit lexical terms that could signal an argumentative relation with the prior turn. Instead, the argumentative move is being implicitly expressed using sarcasm. 



We study whether modeling \emph{sarcasm} can improve the detection and classification of \emph{argumentative relations} in online discussions. 
We propose a thorough experimental setup to answer this question using feature-based machine learning approaches and deep learning models. For the former, we show that \emph{combining} features that are useful to detect sarcasm \cite{joshi2015,muresanjasist2016,ghosh2018marker} with state-of-the-art argument features leads to better performance for the argumentative relation classification task (agree/disagree/none) (Section \ref{section:results}). For the deep learning approaches, we hypothesize that \emph{multitask learning}, which allows representations to be shared between multiple tasks (e.g., here, the tasks of argumentative relation classification and sarcasm detection),  lead to better generalizations. We investigate the impact of multitask learning for a dual Long Short-Term Memory (LSTM) Network with hierarchical attention \cite{ghosh2017role} (Section \ref{subsection:lstm}) and BERT (Bidirectional Encoder Representations from Transformers) \cite{devlin2018bert}, including an optional joint multitask learning objective with uncertainty-based weighting of task-specific losses \cite{kendall2018multi} (Section \ref{subsection:transformer}). We demonstrate that multitask learning improves the performance of the  argumentative relation classification task for all settings (Section \ref{section:results}). 
We provide a detailed qualitative analysis (Section \ref{subsection:qual}) to give insights into when and how modeling sarcasm helps. 
We make the code from our experiments publicly available.\footnote{https://github.com/ritvikshrivastava/multitask\_transformers} The Internet Argument Corpus ($IAC$) \cite{walker2012corpus} can be found for public acess here:\footnote{https://nlds.soe.ucsc.edu/iac2}

\section{Related Work} \label{section:related}

Argument mining is a growing area of research in computational linguistics, focusing on the detection of argumentative structures in a text (see \newcite{stede2018argumentation} for an overview). 
This paper focuses on two subtasks: argumentative relation identification and classification (i.e., agree/disagree/none). 
Some of the earlier work on argumentative relation identification and classification has relied on feature-based machine learning models, focusing  on online discussions  \cite{abbott2011,walker2012corpus,misra2013,ghosh2014analyzing,wacholder2014annotating} and monologues \cite{stab2014identifying,stab-gurevych-2017-parsing,persing-ng-2016-modeling,ghosh-etal-2016-coarse}.  
\newcite{stab2014identifying} proposed a set of lexical, syntactic, semantic, and discourse features to classify them. On the same essay dataset, \newcite{nguyen2016context} utilized contextual information to improve the accuracy. Both \newcite{stab-gurevych-2017-parsing} and \newcite{persing-ng-2016-modeling} used Integer Linear Programming (ILP) based joint modeling to detect argument components and relations. \newcite{rosenthal2015couldn} introduced sentence similarity and accommodation features, whereas \newcite{menini2016agreement} presented how entailment between text pairs can discover argumentative relations. Our argumentative features in the feature-based model are based on the above works (Section \ref{subsection:lr}). We show that additional features that are useful in sarcasm detection \cite{joshi2015, ghosh2018marker} enhance the performance on the argumentative relation identification and classification tasks.

In addition to feature-based models, deep learning models have been recently used for these tasks.  \newcite{potash2016here} proposed a pointer network, and \newcite{hou-jochim-2017-argument} offered LSTM+Attention network to predict argument components and relations jointly, whereas \cite{chakrabarty-etal-2019-ampersand} exploited adaptive pretraining \cite{gururangan2020dont} for BERT to identify argument relations. We use two multitask learning objectives (argumentative relation identification/classification and sarcasm detection), as our goal is to investigate whether identifying sarcasm can help in modeling the disagreement space. \newcite{majumder2019sentiment,chauhan2020sentiment} used multitask learning for sarcasm \& sentiment and sarcasm, sentiment, \& emotion, respectively, where a direct link between the corresponding tasks is evident.    

Finally, analyzing the role of sarcasm and verbal irony in argumentation has a long history in linguistics \cite{tindale1987use, gibbs2005irony,Averbeck2013,vanEemren1992}. 
We propose joint modeling of argumentative relation detection and sarcasm detection to empirically validate sarcasm's role in shaping the disagreement space in online conversations. 

While the focus of our paper is not to provide a state-of-the-art sarcasm detection model, 
our feature-based models, along with the deep learning models for sarcasm detection are based on state-of-the-art approaches. We implemented discrete features such as pragmatic features \cite{gonzalez-ibanez-etal-2011-identifying,muresanjasist2016}, diverse sarcasm markers \cite{ghosh2018marker}, and incongruity detection features \cite{riloff,joshi2015}. The LSTM models are influenced by \newcite{ghosh2017magnets,ghosh2018sarcasm}, where the function of contextual knowledge is used to detect sarcasm. Lastly, transformer models such as BERT and RoBERTa have been used in the winning entries for the recent shared task on sarcasm detection \cite{ghosh-etal-2020-report}. In our research, for both kinds of deep-learning models, the best results are obtained by using the multitask setup, showing that multitask learning indeed helps improve both tasks.

\section{Data} \label{section:data}





Our $training$ and $test$ data are collected from the Internet Argument Corpus ($IAC$) \cite{walker2012stance}. This corpus consists of posts from conversations in online forums on a range of controversial political and social topics 
such as Evolution, Abortion, Gun Control, and Gay Marriage \cite{abbott2011,abbott2016internet}. 
Multiple versions of $IAC$ corpora are publicly available, and we use a particular subset, marked as $IAC_{orig}$, collected from \newcite{abbott2011}. This consists of around 10K pairs of conversation turns (i.e., prior turn $pt$ and the current turn $ct$)  that were annotated using Mechanical Turk for argumentative relations (agree/disagree/none) and other characteristics such as sarcasm/non-sarcasm, respect/insult, nice/nastiness. Median Cohen's $\kappa$ is 0.5 across all topics.


\begin{table}
\centering
\begin{tabular}{ccc}
\hline
Arg. Rel. & Sarcasm & \# of turns \\
\hline
\multirow{2}{*}{$A$} &  $S$ & 315 (33\%)  \\
& $NS$ & 638 (67\%)\\
\hline
\multirow{2}{*}{$D$} &  $S$ & 2207 (57\%) \\
& $NS$ & 1696 (43\%) \\
\hline
\multirow{2}{*}{$N$} &  $S$ & 2285 (44\%) \\
& $NS$ & 2841 (56\%) \\
\hline
\end{tabular}
\caption{Dataset statistics; A (Agree), D (Disagree), N (None); S (Sarcasm), NS (Non-Sarcasm)}
\centering
\label{table:argusarctable}
\end{table}


 For agree/disagree/none relations the annotation was a 
  scalar judgment on an 11 point scale [-5,5] where ``-5'' indicates a  high disagreement move, ``0'' indicates none relation, and ``5'' denotes a high agreement move. 
We converted the scalar values to three categories: disagree ($D$) for values between [-5, -2], none ($N$) for values between [-1,1],  and agree ($A$) for values between [2,5], where the scalar partitions ([]) follow prior work with $IAC$ \cite{misra2013,rosenthal2015couldn}.

Each ``current turn" that is part of a $<$pt$, $ct$>$ pair is also labeled with a Sarcasm ($S$) or Non-Sarcasm ($NS$) label. Table \ref{table:argusarctable} shows the data statistics in terms of argumentative relations ($A$/$D$/$N$) and sarcasm ($S$/$NS$). We split the dataset into $training$ (80\%; 7,982 turn pairs), $test$ (10\%; 999 turn pairs), and $dev$ (10\%; 999 turn pairs) sets where each set contains a proportional number of instances (i.e., 80\% of 315 (=252) sarcastic turns ($S$) with argument relation label $A$ (agree) appears in the training set). The $dev$ set is used for parameter tuning.

\section{Experimental Setup} \label{section:method}

We present the computational approaches to investigate whether modeling \emph{sarcasm} can help detect argumentative relations. As our goal is to provide a comprehensive empirical investigation of sarcasm's role in argument mining rather than propose new models, we explore three separate machine learning approaches well-established for studying argumentation and figurative language. First, we implement a Logistic Regression method that exploits a combination of state-of-the-art features to detect argumentative relations as well as sarcasm (Section \ref{subsection:lr}). Second, we present a dual LSTM architecture with hierarchical attention and its multitask learning setup (Section \ref{subsection:lstm}). Third, we  discuss experiments using the pre-trained BERT models and our multitask learning architectures based on it (Section \ref{subsection:transformer}).  

\subsection{Logistic Regression with Discrete Features} \label{subsection:lr}

We use a Logistic Regression (LR) model that uses both argument-relevant ($ArgF$) and sarcasm-relevant ($SarcF$) features. Unless mentioned, all features were extracted from the current turn $ct$. 
\paragraph{Argument-relevant features ($ArgF$).} We first evaluate the features that are reported as being useful for identifying and classifying argumentative relations: (a) \emph{n-grams} (e.g., unigram, bigram, trigram) created based on the full vocabulary of the $IAC$ corpus; (b) \emph{argument lexicons}: two lists of twenty words representing agreement (e.g., ``agree'', ``accord'') and disagreement (e.g., ``differ'', ``oppose''), respectively \cite{rosenthal2015couldn} (c) 
\emph{sentiment lexicons} such as MPQA \cite{wilson2005recognizing} and {opinion lexicon} \cite{hu2004} to identify sentiment in the turns; (d) \emph{hedge features}, since they are often used to mitigate speaker's commitment \cite{tanetal16}; (e) \emph{PDTB discourse markers} because \emph{claims} often start with discourse markers such as \emph{therefore}, \emph{so}. We discard markers from the temporal relation; (f) \emph{modal verbs} because they signal
the degree of certainty when expressing a claim \cite{stab2014identifying}; (g) \emph{pronouns}, since they dialogically point to the previous speaker's stance; 
(h) \emph{textual entailment}: captures whether a position expressed in the prior turn is accepted in the current turn \cite{cabrio2012combining,menini2016agreement}\footnote{We used the textual entailment toolkit (AllenNLP) \cite{Gardner2017AllenNLP}.}; (i) \emph{lemma overlap} to determine topical alignment between the prior and current turn \cite{somasundaran2010recognizing}. We compute lemma overlap of noun, verbs, and adjectives between the turns, and (j) \emph{negation} to extract explicit negation cues (e.g., ``not'', ``don't'') that often signal disagreement.   
\paragraph{Sarcasm-relevant features ($SarcF$).}  As sarcasm-relevant features we use: 
(a) Linguistic Inquiry Word Count \emph{(LIWC)} \cite{pennebaker2001} features to capture the linguistic, social, individual, and psychological processes; (b) measuring \emph{sentiment incongruity}, that is, capturing the number of times the difference in sentiment polarity between the prior turn $pt$ and the current turn $ct$ occurs and number of positive and negative sentiment words in turns \cite{joshi2015}; (c) \emph{sarcasm markers} used by \newcite{ghosh2018marker}, such as \emph{capitalization}, \emph{quotation marks}, \emph{punctuation}, \emph{exclamations} that emphasize a sense of surprisal, \emph{tag questions}, \emph{interjections} because they seem to undermine a literal evaluation, \emph{hyperbole} because users frequently overstate the magnitude of an event in sarcasm, and \emph{emoticons} \& \emph{emojis}, since they often emphasize the sarcastic intent.

We use \emph{SKLL}, an open-source Python package that wraps around the Scikit-learn tool \cite{pedregosa2011scikit}. \footnote{https://pypi.org/project/skll/} We perform the feature-based experiment using the Logistic Regression model from Scikit-learn. 

In the experimental runs,  LR$_{ArgF}$ (i.e., model that uses just the $ArgF$ features) denotes the \emph{individual} model and LR$_{ArgF+SarcF}$ (i.e., model that uses both $ArgF$ and $SarcF$ features) is the \emph{joint} model.

\subsection{Dual LSTM and Multitask Learning} \label{subsection:lstm}

LSTMs are able to learn long-term dependencies \cite{hochreiter1997long} and have been shown to be effective in Natural Language Inference (NLI) research, where the task is to establish the \emph{relationship} between multiple inputs \cite{rocktaschel2015reasoning}. This type of architecture is often denoted as the \emph{dual architecture} since one LSTM models the premise and the other models the hypothesis (in Recognizing Textual Entailment(RTE) tasks).  \newcite{ghosh2018sarcasm} used the dual LSTM architecture with hierarchical attention (HAN) \cite{yang2016hierarchical} for sarcasm detection to model the conversation context, and we use their approach in this paper to model the current turn $ct$ and the prior turn $pt$. HAN implements attention both at the word level and sentence level. The distinct characteristics of this attention is that the word/sentence-representations are weighted by measuring similarity with a word/sentence level context vector, respectively, which are randomly initialized and jointly learned during training \cite{yang2016hierarchical}.
 We compute the vector representation for the current turn $ct$ and prior turn $pt$ and concatenate vectors from the two LSTMs for the final softmax decision (i.e., $A$, $D$ or $N$ for argumentative relation detection). Henceforth, this dual LSTM architecture is denoted as $LSTM_{attn}$.

\begin{figure}[t]
\centering
\includegraphics[width=7.5cm]{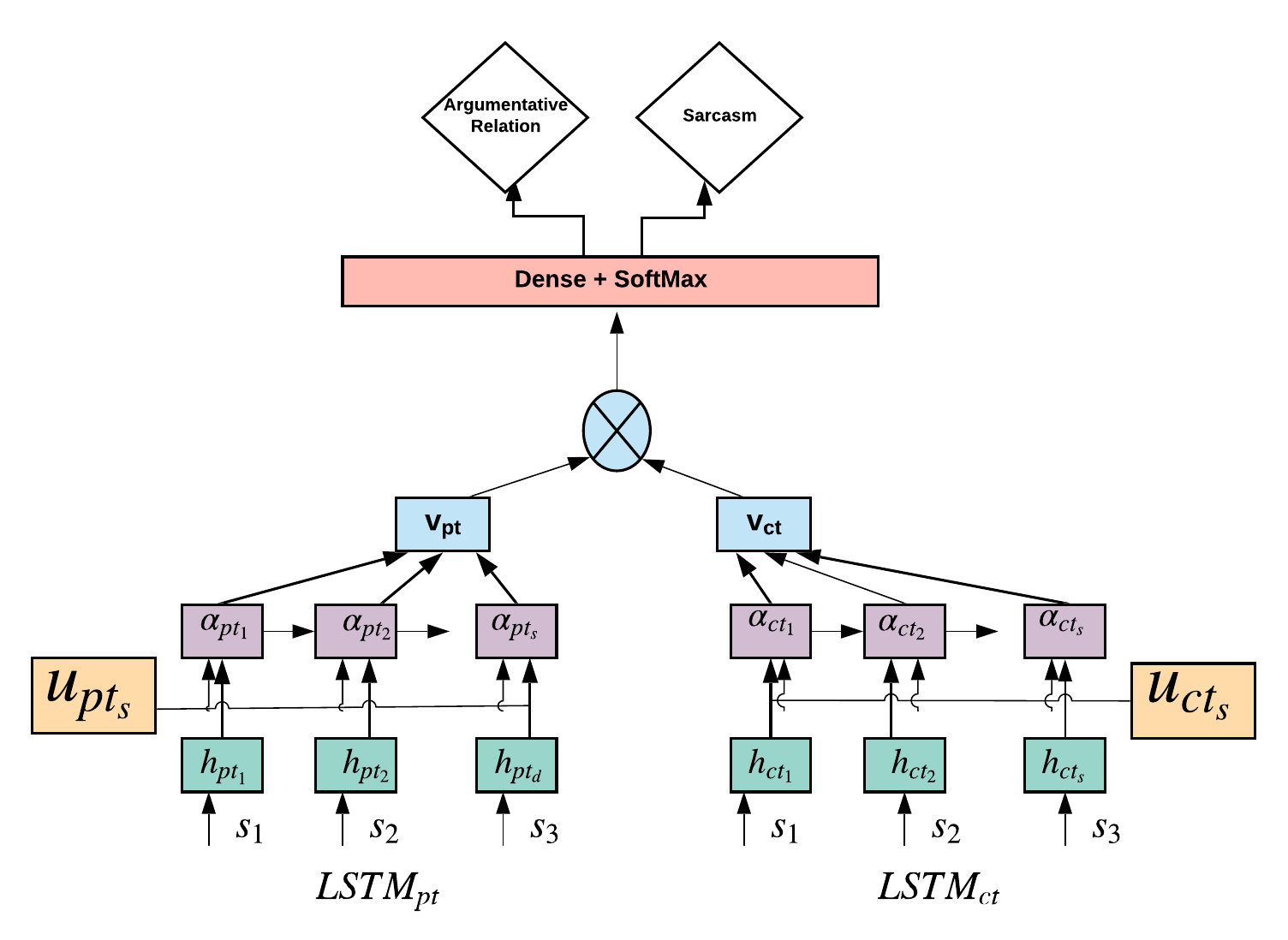}

\caption{Sentence-level Multitask Attention Network for prior turn $pt$ and current turn $ct$. Figure is inspired by \citet{yang2016hierarchical}.}
\label{figure:model}
\end{figure}


To measure the impact of \emph{sarcasm} in argumentative relation detection, we use a multitask learning approach. Multitask learning aims to leverage useful information in multiple related tasks to improve each task's performance \cite{caruana1997multitask,liu2019multi}. We use a simple hard parameter sharing network. The architecture is a replica of the $LSTM_{attn}$, with a modification of employing two loss functions, one for sarcasm detection (i.e., training using the $S$ and $NS$ labels) and another for the argumentative relation classification task (i.e., training using the $A$, $D$, and $N$ labels).      

Figure \ref{figure:model} shows the high-level architecture of the dual LSTM and multitask learning ($LSTM_{MT}$). The prior turn $pt$ (left) and the current turn $ct$ (right) are read by two separate LSTMs (i.e., $LSTM_{pt}$ and $LSTM_{ct}$). In case of $LSTM_{MT}$ the concatenation of $v_{pt}$ and $v_{ct}$ is passed through a dense+Softmax layer for the MTL as shown in Figure \ref{figure:model}. Similar to the $LR$ models,  $LSTM_{attn}$ now represents the \emph{individual} model (i.e., predicts only the argumentative relation) whereas $LSTM_{MT}$ represents the \emph{joint} model.

\paragraph{Dynamic Multitask Loss.}
In addition to simply adding the two losses, we also employed \emph{dynamic weighting} of task-specific losses during the training process, based on the homoscedastic uncertainty of tasks, as proposed in \newcite{kendall2018multi}:
\begin{equation}
L =\sum_{t} \frac{1}{2\sigma^2_{t}}L_{t} + \log\sigma^2_{t}
\end{equation}
where $L_{t}$ and $\sigma_{t}$ depict the task-specific loss and its variance, respectively, over training instances. We denote this variation as LSTM$_{{MT}_{uncert}}$.

\subsection{Pretrained BERT and Multitask Learning} \label{subsection:transformer}

BERT \cite{devlin2018bert}, a bidirectional transformer model, has achieved state-of-the-art performance for many NLP tasks. BERT is initially trained on masked token prediction and next sentence prediction tasks over  large corpora (English Wikipedia and Book Corpus). During its training, a special token ``[CLS]'' is added to the beginning of each training instance, and  the ``[SEP]'' tokens are  added to indicate the end of utterance(s) and separate, in case of two utterances (e.g., $pt$ and $ct$). During the evaluation, the learned representation for the ``[CLS]'' token is processed by an additional layer with nonlinear activation. In its standard form, pre-trained BERT (``bert-base-uncased'') can be used for transfer learning by fine-tuning on a downstream task, i.e., argument relation detection where training instances are labeled as $A$, $D$, and $N$. We denote the BERT baseline model as $BERT_{orig}$ that is fine-tuned over the $training$ partition of only the argumentative relation data (i.e., individual task training). Unless mentioned otherwise, we use the BERT predictions available via the ``[CLS]'' token. To this end, we propose a couple of variations in the multitask learning settings, and they are briefly described in the following sections.

\begin{figure}[t]
\centering
\includegraphics[width=7.5cm]{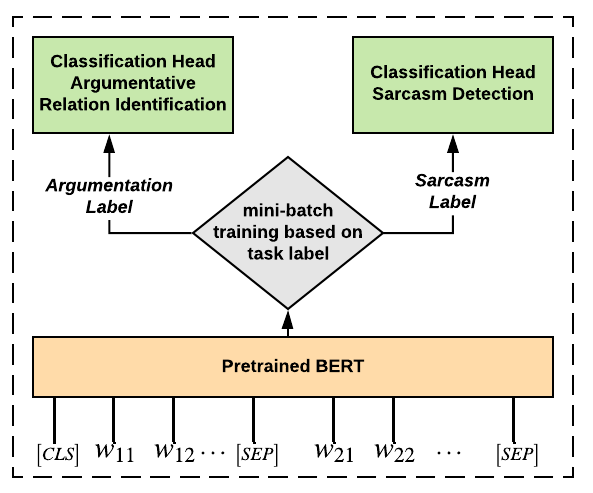}

\caption{Alternating mini-batch training based on the task type ($BERT_{ALT}$).}
\label{figure:bertalt}
\end{figure}

\paragraph{Multitask Learning with BERT.}
The first model we use for multitask learning is denoted as $BERT_{MT}$ (i.e., BERT Multitask Learning). Here, we pass the BERT output embeddings to two classification heads - one for each task (i.e., detection of argumentative relation and sarcasm), and the relevant gold labels are passed to them. Each classification head is a linear layer (size=3 and 2 for \# of labels for argumentative relation and sarcasm detection, respectively) applied on top of the pooled BERT output. The losses from these individual heads are added and propagated back through the model. This allows BERT to model the nuances of both tasks and their interdependence simultaneously.

\noindent\textbf{Dynamic Loss}: Similar to the LSTM architecture, here, too, we experiment with dynamic multitask loss. We denote this variation as BERT$_{{MT}_{uncert}}$. 

\paragraph{Alternate Multitask Learning.}
We employ another multitask learning technique  where we attempt to enrich the learning with fine-tuning of labeled \emph{additional} material from the sarcasm detection task. Notably, we exploit ``sarcasm V2'', a sarcasm detection dataset that was also curated from the original corpus of $IAC$ and was released by \newcite{oraby2016creating}. We pre-process the ``sarcasm V2'' dataset by removing duplicates that appear in $IAC_{orig}$ and we end up selecting 3513 $training_{v2}$ instances and 423 $dev_{v2}$ instances balanced between S/NS categories for experiments and merged them to the sarcasm dataset ($training$ and $dev$, respectively) from $IAC_{orig}$. Note, unlike the original multitask setting, this time we have more sarcastic instances (a total of 11,495) than instances labeled with argumentative roles (7,982 instances as before) for the training purpose, while keeping the $test$ set from $IAC_{orig}$ unchanged. 

Since the $training$ data is now unequal between the two tasks of argumentative relation and sarcasm detection, we create mini-batches so that each batch consists of instances with only one task label (i.e., either argumentative labels or sarcasm labels). The batches from the two tasks are interleaved uniformly, i.e., the BERT model is only passed to one of the two tasks' specific classification heads, and the related loss is used to update the parameters in that iteration. This way, the model trains both tasks but alternates between the two tasks per mini-batch iteration while the extra batches of sarcasm data from the ``sarcasm V2'' dataset are managed at the end together. This model is denoted as $BERT_{ALT}$ (see Figure \ref{figure:bertalt}). 



For brevity, all models' parameter tuning description (e.g., Logistic Regression, Dual LSTM, BERT) is in the supplemental material.



\section{Results and Discussion} \label{section:results}



 \begin{table}[]
 \center
 \begin{tabular}{cccccc}
 \hline
 Model  & $F1_{micro}$ & $A$ & $D$ & $N$ \\
 \hline
 {LR$_{ArgF}$} & 53.5 & 22.4  & 57.2 & 56.3  \\
 {LR$_{ArgF+SarcF}$}  & \textbf{56.4}$^{\alpha{^{*}}}$  &	31.0 & {58.4} & {58.9}\\
\hline
{LSTM$_{Attn}$} & 51.8 & 28.0 & 49.4 & 59.2\\
{LSTM$_{MT}$} & 53.1  & 30.0 & 53.2 & 56.5 \\
{LSTM$_{{MT}_{uncert}}$}  & \textbf{54.6} $^{\alpha{^{*}}}$ & 33.1 & 54.5 & 58.5 \\

 \hline
 {BERT$_{orig}$}  & 62.2 &	41.8 &	63.3 & 64.4\\
{BERT$_{MT}$} & 63.2 & {44.5} & {64.1} & 65.4 \\
{BERT$_{{MT}_{uncert}}$} & \textbf{65.3}$^{\alpha{^{*}}}$ & {44.6} & {66.2} & {67.5} \\
\hline
{BERT$_{ALT}$} & 63.4 & 40.1 & 62.2 & {66.9} \\

\hline
\end{tabular}
 \caption{Results for argumentative relation detection ($F1_{micro}$ and F1 scores/category) on the $test$ set of $IAC_{orig}$. $^{\alpha{^{*}}}$ depict significance on $p\le 0.05$ (measured via Mcnemar's test) against the corresponding individual model (e.g., LR$_{ArgF}$, LSTM$_{Attn}$, BERT$_{orig}$, respectively). Highest scores per group of models are in \textbf{bold}.}
 \label{table:allargresults}
 \end{table}

Table \ref{table:allargresults} presents the classification results on the $test$ set. We report F1 scores for each class ($A$, $D$ and $N$) and Micro-F1 overall score (F1$_{micro}$) (used to account for multi-class and class imbalance). 

\textbf{The LR model} using both the $SarcF$ and $ArgF$ features performs better than the model that uses $ArgF$ features alone, improving the overall performance by an absolute 2.9\% F1$_{micro}$, and showing a huge  impact on the agreement class ($A$) (8.6\% absolute improvement). Table \ref{table:topargfeatures} shows the \emph{top} discrete features for argumentative relation identification. From $ArgF$ features (first column), we notice discourse expansion (``particularly''), contrast (``although'') and agree/disagree lexicon getting high feature weights. We also notice \emph{pronouns} receive large feature weights because argumentative text often refers to personal stance (e.g., ``you think'', ``I believe''). However, when analyzing ${ArgF+SarcF}$ features we find various sarcasm markers, such as tag questions, hyperbole, multiple punctuation, or sarcasm characteristics such as sentiment incongruity receive the highest weights. 


\begin{table}[t]
\centering
\begin{small}
\begin{tabular}{ p{3.5cm}p{ 3.5cm} }
\hline
LR$_{ArgF}$ & LR$_{ArgF+SarcF}$ \\
\hline
\emph{pronouns}: I. my (both $A$), your(s) ($D$); \emph{discourse}: so, because, for (all $A$), incidentally, particularly, although (all $D$); \emph{disagree\_lexicon}: disagree, differ (both $D$);\emph{agree\_lexicon}: agreed ($A$); \emph{entailment relation}; \emph{negation} ($D$) & 
\emph{pronouns}: mine, my (both $A$), you ($D$); \emph{discourse}: then ($A$), though, however (both $D$); \emph{modal}: will ($A$); \emph{punctuation}: multiple question marks (both $A$ and $D$); \emph{tag question}: ``are you'', ``do you'' (both $D$); \emph{hyperbole}: wonderful ($A$), nonsense, biased (both $D$); \emph{LIWC dimensions}: anxiety, assent, certainty (all $D$); \emph{sentiment incongruity} ($D$); \emph{interj}: so, agreed (both $A$)  \\
\hline
\end{tabular}
\end{small}

\caption{Top discrete features from  LR$_{ArgF}$ and LR$_{ArgF+SarcF}$ models, respectively. $A$ and $D$ depict the argumentative relations (agree and disagree) for the particular feature.} 
\centering
\label{table:topargfeatures}
\end{table}

For \textbf{LSTM models}, we see that multitask learning helps, {LSTM$_{{MT}_{uncert}}$} showing a 2.8\% improvement over the single model LSTM$_{Attn}$, which is statistically significant. Moreover, we notice that the improvement for the agree ($A$) and disagree ($D$) classes is 5.1\%, with just a small reduction for the none ($N$) class (0.7\%).  

For \textbf{BERT}, we  notice better results when performing multitask learning, while the best performing model is obtained from BERT$_{{MT}_{uncert}}$ where we  experimented with the dynamic weighting of task-specific losses during the training process \cite{kendall2018multi}. The performance increase is consistent across all three classes. The difference in performance among each setup is statistically significant, as shown in Table \ref{table:allargresults}. Moreover, BERT$_{{MT}_{uncert}}$ model improves the $F1_{micro}$ by a large margin when compared to the LR and the LSTM models. 
However, adding more data for the auxiliary task (i.e., sarcasm detection) as presented in $BERT_{ALT}$ did not  provide any significant improvement, only a 0.2 improvement of $F1_{micro}$ over $BERT_{MT}$ (however it does show improvement over the single task model). 
The reason could be that although
``sarcasm V2''is a subset of the original $IAC$ corpus, it was annotated by a different set of Turkers than $IAC_{orig}$ with different annotation guidelines.

Between the three classes - $A$, $D$, and $N$ - we observe the lowest performance on the $A$ class. 
This is unsurprising, given the highly unbalanced setting of the $training$ data ($A$ occurs less than 10\% of times in the $IAC_{orig}$, see Table \ref{table:argusarctable}). 

In sum, these improvements through multitask learning over single task argumentative relation detection indicate that modeling sarcasm is useful in modeling the disagreement space in online discussions. This provides an empirical justification to existing theories that study sarcasm's impact in modeling argumentation, persuasion, and argument fallacies such as ad hominem attacks. 
Finally, we notice that multitask learning also improves the performance on the sarcasm detection task (results are presented in the Appendix).

\subsection{Qualitative Analysis} \label{subsection:qual}
\begin{figure*}[h] 
\centering 
\begin{framed}
\includegraphics[width=7.5cm]{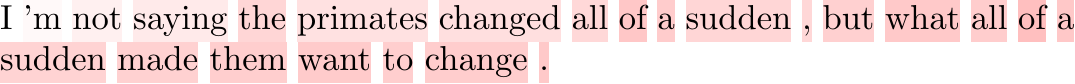} 
\hspace {0.05in} 
\includegraphics[width=7.5cm]{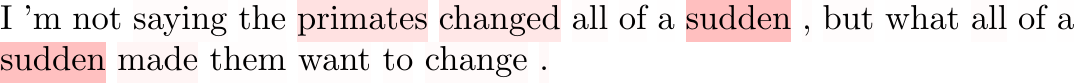}
\vspace{0.05in}

\includegraphics[width=7.5cm]{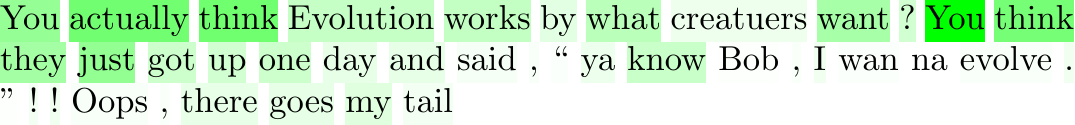}
\hspace {0.05in} 
\includegraphics[width=7.5cm]{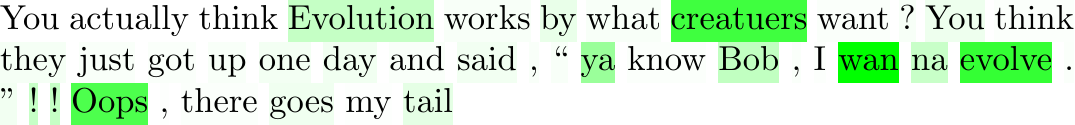}
\end{framed}
 \centering
\caption{Attention heatmap of a particular turn pair  from $LSTM_{attn}$\emph(left) and LSTM$_{{MT}_{uncert}}$\emph(right) showing higher weights on sarcasm marker such as ``Oops'' and ``!!'' for LSTM$_{{MT}_{uncert}}$ (disagree relation)}
\label{figure:lstmcomp1}
\end{figure*}

To further investigate the effect of multitask learning, we present qualitative analysis studies to:
\begin{enumerate}
    \item Understand the models' performance by looking at the turns correctly classified by the multitask models and misclassified by the corresponding individual single task model. We analyze the turns in terms of sarcastic characteristics - whether they depict incongruity, humor, or sarcasm indicators (i.e., markers). 
    \item Understand when both multitask and individual model made incorrect predictions. 
\end{enumerate}

\begin{figure}[t] 
\centering
\begin{framed}
\includegraphics[width=3.3cm]{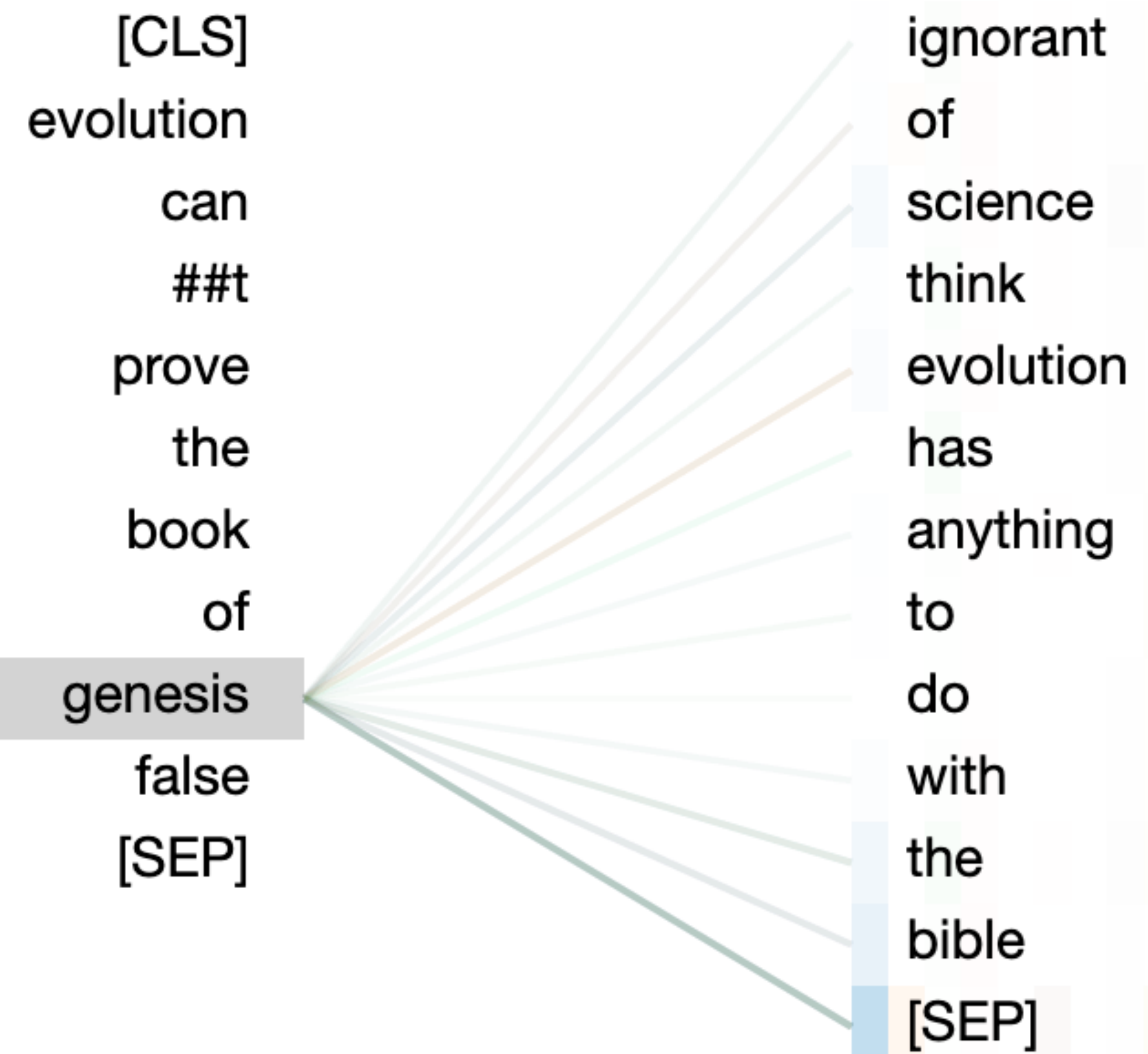}
\hspace {0.05in} 
\includegraphics[width=3.3cm]{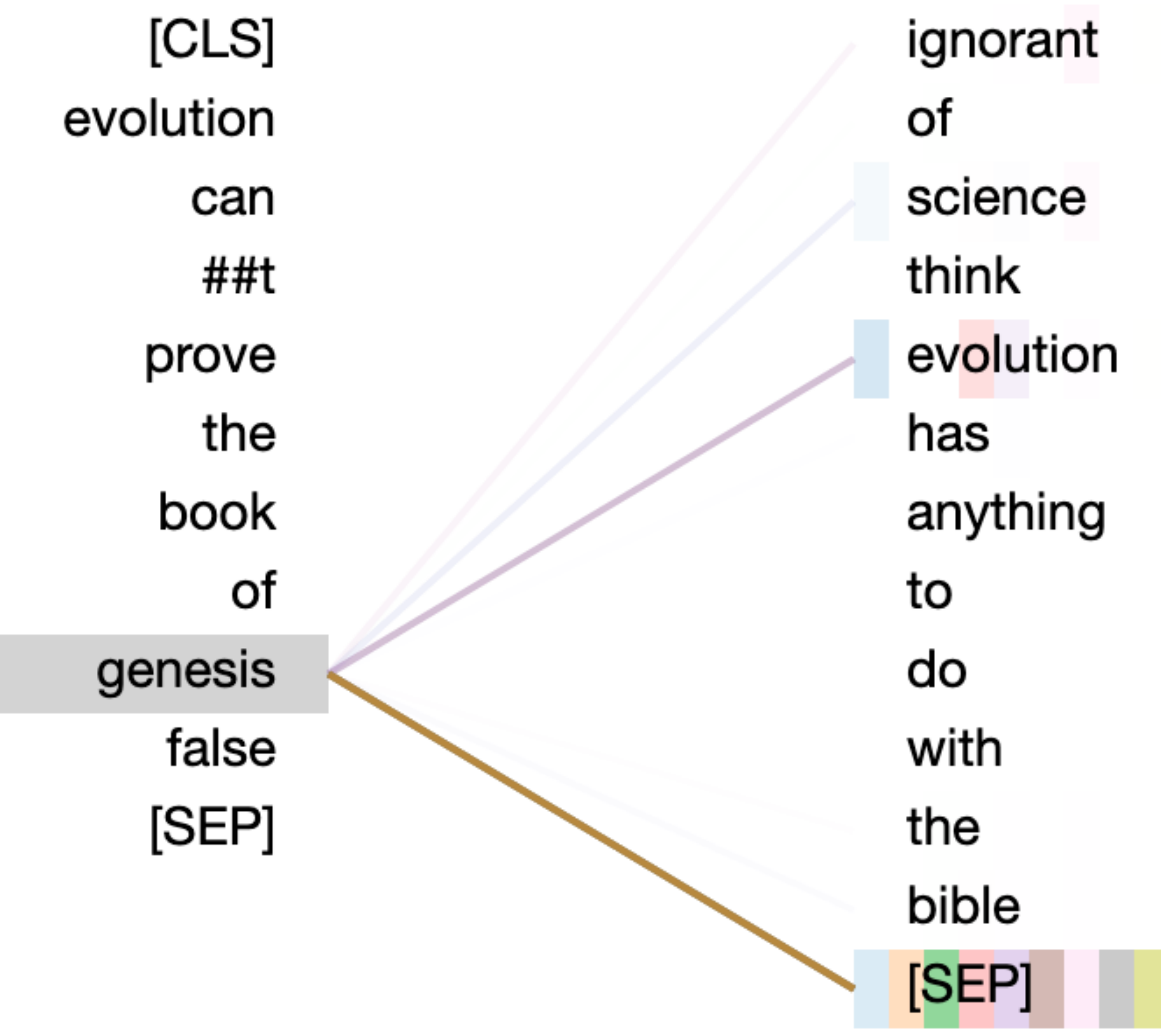}
\end{framed}
 \centering
\caption{$BERT_{{MT}_{uncert}}$ (right) attending contrasting words more in word-level attention in comparison to $BERT_{Orig}$ (left) (disagree relation)}
\label{figure:bertcomp1}
\end{figure}
\begin{figure}[t] 
\centering
\begin{framed}
\includegraphics[width=3.3cm]{bert_single_genesis.pdf}
\hspace {0.05in} 
\includegraphics[width=3.3cm]{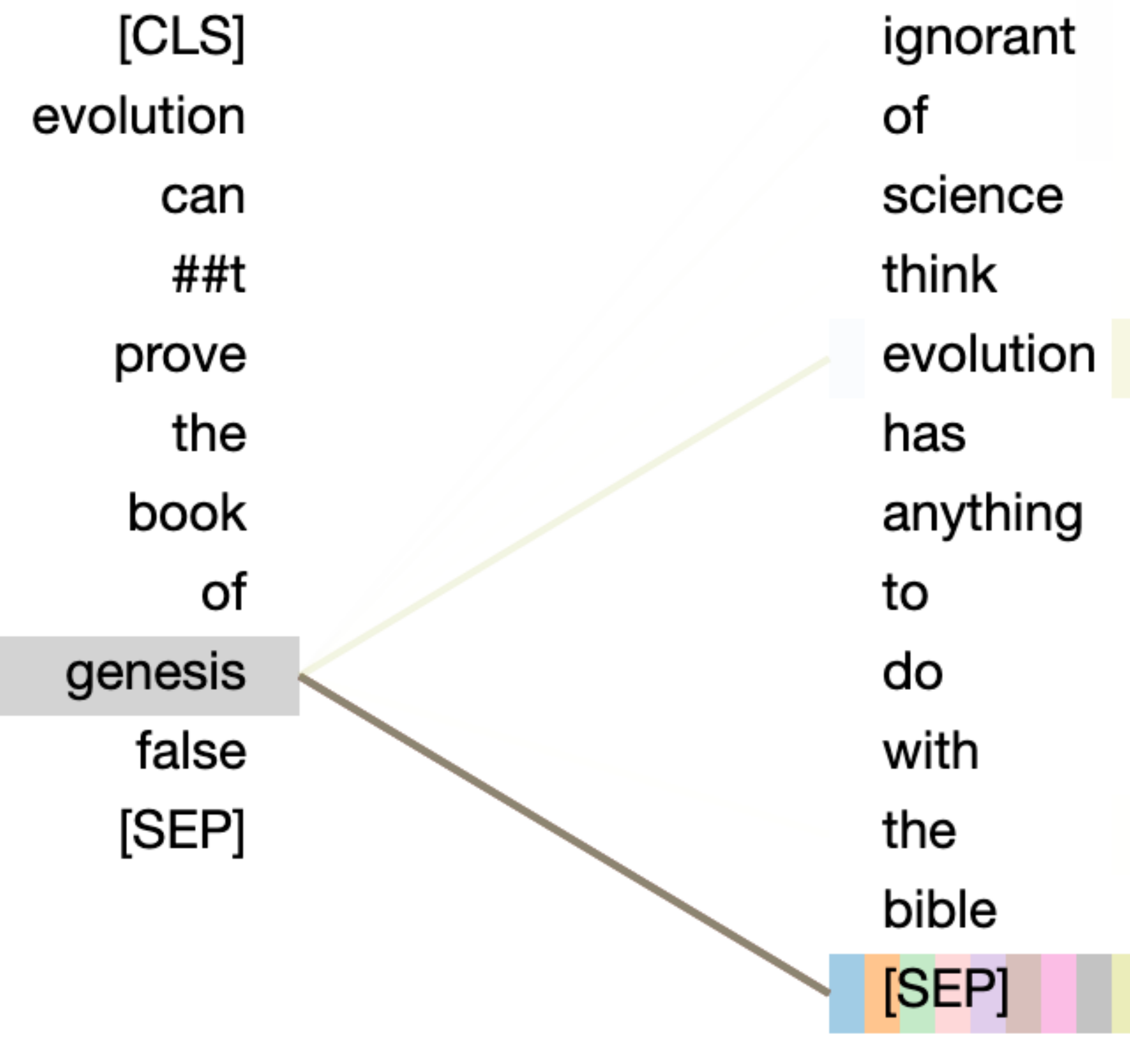}
\end{framed}
 \centering
\caption{$BERT_{ALT}$ (right) attending only contrasting words in comparison to $BERT_{Orig}$ (left) (disagree relation). However, the strength of the contrast in the case of $BERT_{ALT}$ is lower than $BERT_{{MT}_{uncert}}$ for the same example turns.}
\label{figure:bertcompalt}
\end{figure}
\begin{figure}[t] 
\centering
\begin{framed}
\includegraphics[width=3.3cm]{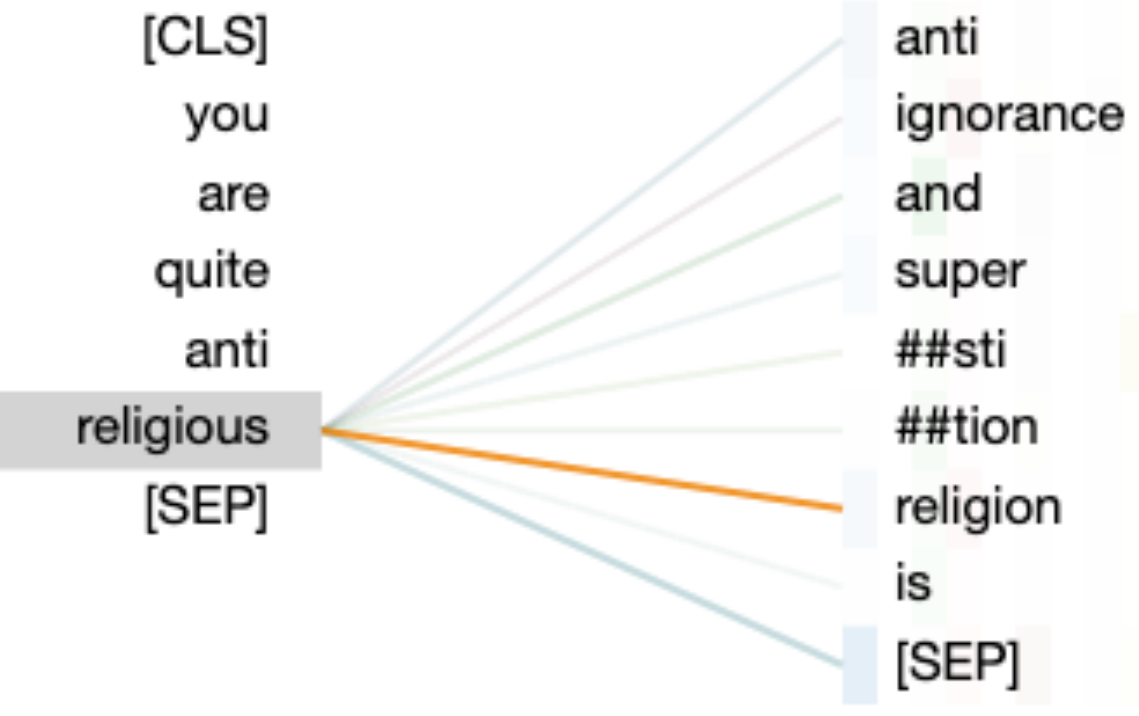}
\hspace {0.05in} 
\includegraphics[width=3.3cm]{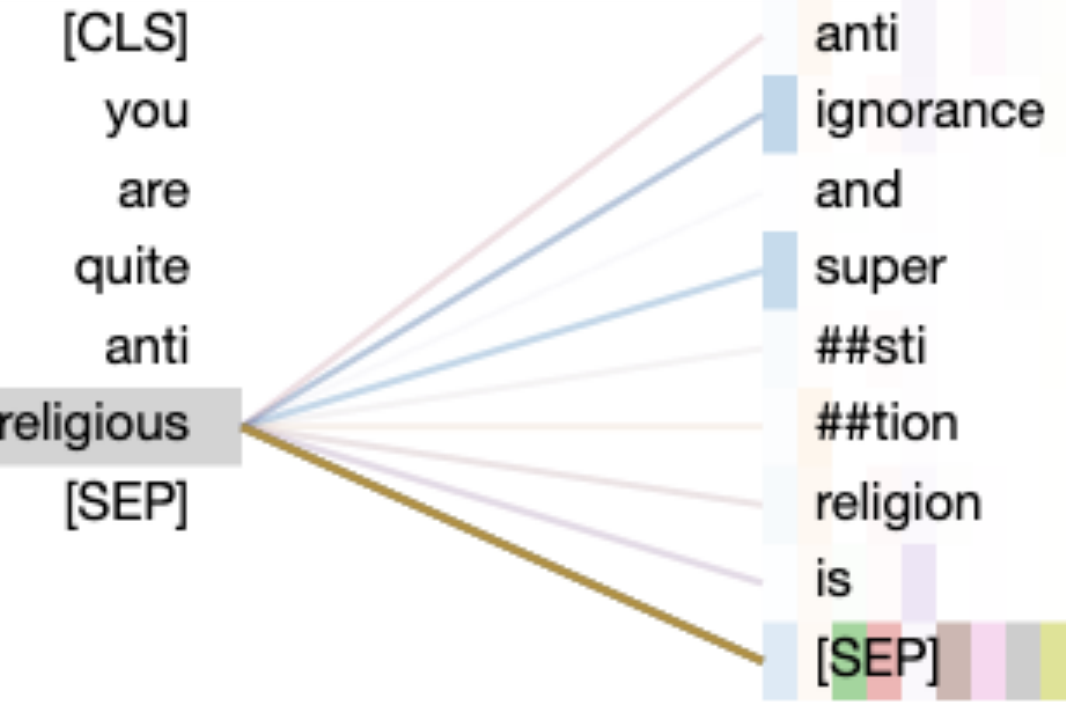}
\end{framed}
 \centering
\caption{$BERT_{{MT}_{uncert}}$ (right) attending contrasting words more than $BERT_{Orig}$ (left) (disagree relation)}
\label{figure:bertcomp2} 
\end{figure}
\begin{figure}[t] 
\centering
\begin{framed}
\includegraphics[width=3.3cm]{bert_single_relig.pdf}
\hspace {0.05in} 
\includegraphics[width=3.3cm]{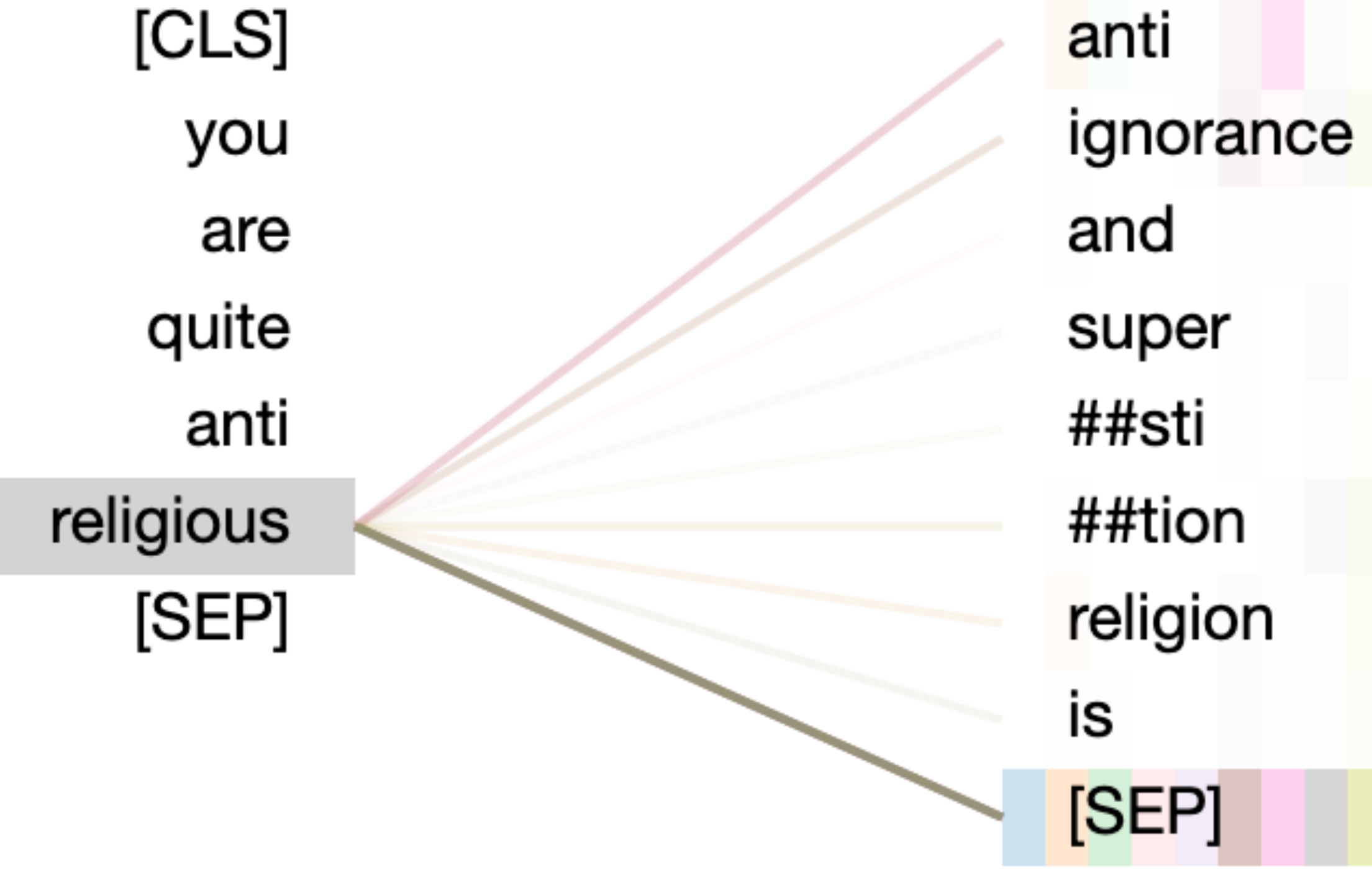}
\end{framed}
 \centering
\caption{$BERT_{ALT}$ (right) attending only the contrasting words in comparison to $BERT_{Orig}$ (left) (disagree relation)}
\label{figure:bertcompalt2} 
\end{figure}
\begin{figure}[t] 
\centering
\begin{framed}
\includegraphics[width=3.3cm]{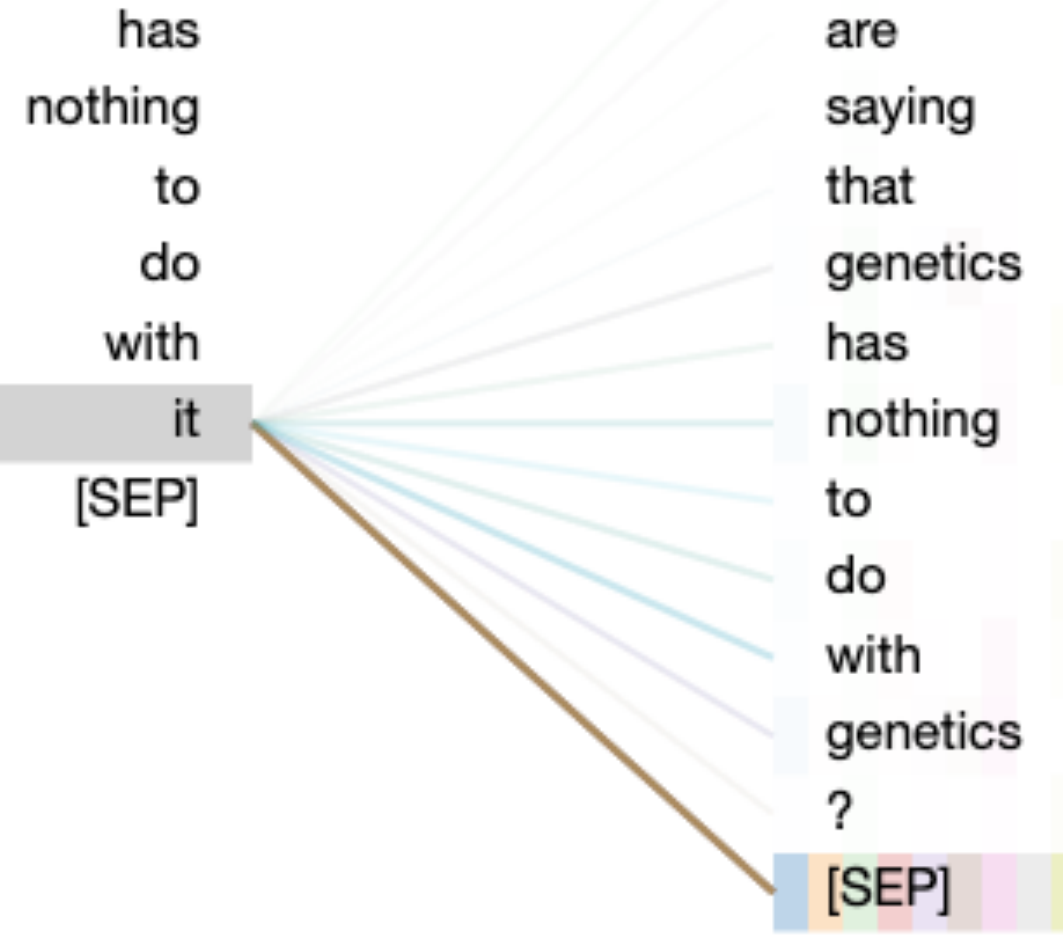}
\hspace {0.05in} 
\includegraphics[width=3.3cm]{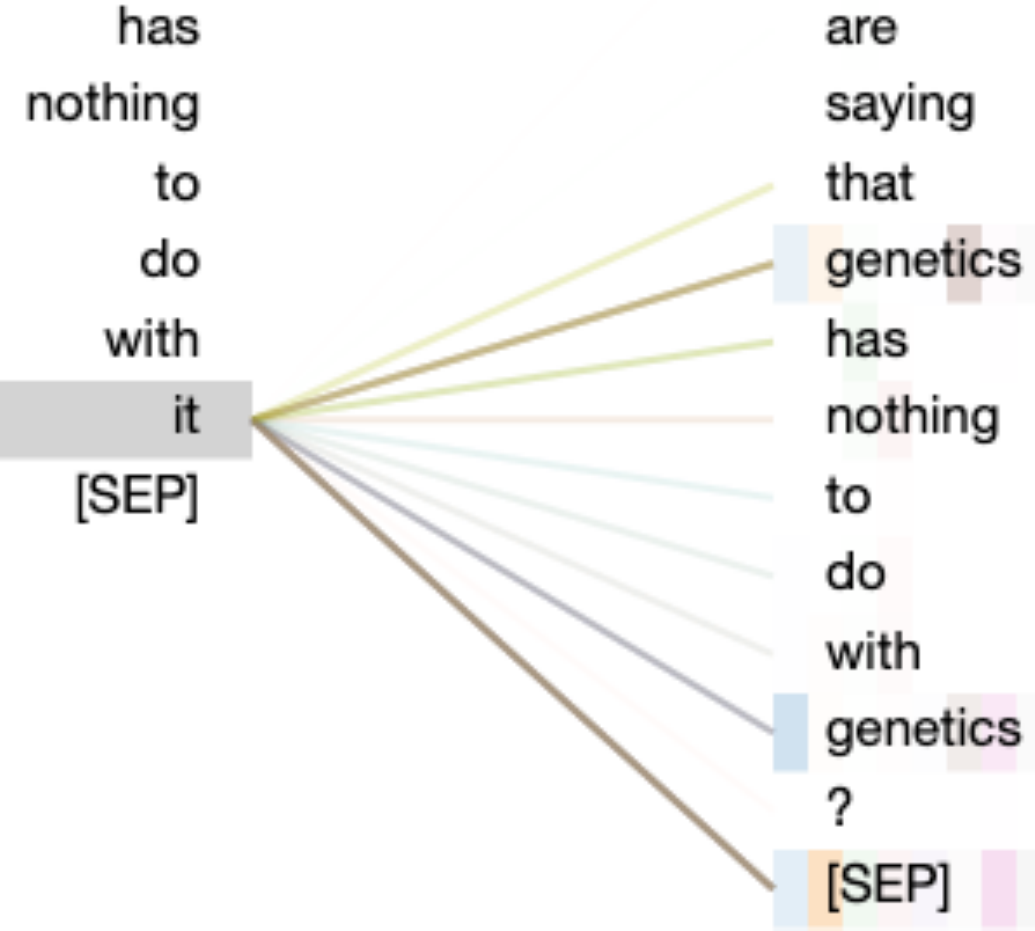}
\end{framed}
 \centering
\caption{$BERT_{{MT}_{uncert}}$ (right) attending co-referenced words in a humorous example missed by the $BERT_{Orig}$ model (left) (disagree relation)}
\label{figure:bertcomp3}
\end{figure}

We compare the predictions between the multitask and the individual models for different settings to address the first issue. For example, $BERT_{{MT}_{uncert}}$ correctly identifies 6 $A$, 50 $D$, and 60 $N$ instances more than $BERT_{Orig}$ (out of 91, 398 and 510 instances, respectively). Two of the authors independently investigated a random sample of 100 instances ($qual$ set) chosen from the union of the $test$ instances that are correctly predicted only by the multitask models (LR$_{ArgF+SarcF}$, $LSTM_{{MT}_{uncert}}$, $BERT_{{MT}_{uncert}}$, and $BERT_{ALT}$) and not by the corresponding individual models (LR$_{ArgF}$, $LSTM_{attn}$, and $BERT_{Orig}$).
For both Transformer and LSTM-based models, we explore how attention heads behave and whether common patterns exist (e.g., attending words with opposite meaning when incongruity occurs). We display the heat maps of the attention weights for a pair of prior and current turns (LSTM-based models) (Figure \ref{figure:lstmcomp1}) whereas for BERT we display word-to-word attentions (Figures \ref{figure:bertcomp1}, \ref{figure:bertcompalt}, \ref{figure:bertcomp2}, \ref{figure:bertcompalt2}, and \ref{figure:bertcomp3}) using visualization tools \cite{vig2019multiscale,yang2018ncrf}.\footnote{\newcite{clark2019does} have probed different layers and attention heads in BERT to find patterns, e.g., whether a token consistently attends a fixed token in a specific layer. To avoid confusion and bias, we select attention examples from only the middle (layer=6) layer.}  All the examples presented in this section are argumentative moves (i.e., turns with $A$ or $D$) correctly identified by our multitask learning models but wrongly predicted as none ($N$) by the individual models. Moreover, the multitask learning models also correctly predict that these turns are instances of sarcasm.

\paragraph{Incongruity between prior turn and current turn.} Semantic incongruity, which can appear between conversation context $pt$ and the current turn $ct$ is an inherent characteristic of sarcasm \cite{joshi2015}. This characteristic highlights the inconsistency between \emph{expectations} and \emph{reality}, making sarcasm or irony highly effective in persuasive communication \cite{gibbs2005irony}. 


In the case of BERT, Figure \ref{figure:bertcomp1} presents the turns ``evolution can't prove the book of genesis false'' ($pt$) $\leftrightarrow$ ``ignorant of science think evolution has anything to do with the bible'' ($ct$). Here, $BERT_{{MT}_{uncert}}$ shows more attention between incongruous terms (``genesis'' $\leftrightarrow$ ``science'', ``evolution'') as well as to the mocking word ``ignorance". Likewise, Figure \ref{figure:bertcomp2}  presents two turns  ``you are quite anti religious it seems'' ($pt$) $\leftrightarrow$ ``anti ignorance and superstition \dots this is religion'' ($ct$).  We notice the word ``religious'' is attending ``anti'' and ``ignorance'' with high weights in case of $BERT_{{MT}_{uncert}}$ (from $pt$ to $ct$) whereas $BERT_{Orig}$ only attends to the word ``religious'' from the $pt$ to $ct$ turn. By modeling sarcasm, the multitask learning models can better predict argumentative moves that are expressed implicitly. 

We also evaluate the $BERT_{ALT}$ model for the examples presented in Figure \ref{figure:bertcomp1} and Figure \ref{figure:bertcomp2}. Figure \ref{figure:bertcompalt} shows that although $BERT_{ALT}$ is attending (from $pt$ to $ct$) incongruous terms ``genesis'' $\leftrightarrow$ ``evolution'', the strength of the relation (i.e., attention weight) is comparatively lower than $BERT_{{MT}_{uncert}}$ (See Figure \ref{figure:bertcomp1}). On the contrary, between Figure  \ref{figure:bertcomp2} and Figure \ref{figure:bertcompalt2}, $BERT_{{MT}_{uncert}}$ model is attending multiple words in $ct$ from the word ``religion'' in $pt$, but the $BERT_{ALT}$ model attends only two words `anti'' and ``ignorance'', with high weights from ``religion'' ($pt$ to $ct$). 

\paragraph{Humor by word repetition.}
Often the current turn $ct$ sarcastically taunts the prior turn $pt$ by word repetition and rhyme, imposing a humorous comic effect, also regarded as the phonetic style of humor \cite{yang2015humor}. For the pair, ``genetics has nothing to do with it'' ($pt$) $\leftrightarrow$ ``are saying that genetics has nothing to do with genetics?'' ($ct$), we notice in $BERT_{{MT}_{uncert}}$ the token ``it'' in $pt$ correctly attends to both occurrences of ``genetics'' in $ct$ where the second occurrence is the co-reference of ``it'' (Figure \ref{figure:bertcomp3}), which is missed by the individual model $BERT_{Orig}$.    

\paragraph{Role of sarcasm markers.} 
Sarcasm markers are indicators that alert if an utterance is sarcastic \cite{attardo2000ironya}. While comparing the logistic regression models between LR$_{ArgF+SarcF}$ and LR$_{ArgF}$, we observe markers such as multiple punctuations (``???''), tag question (``are you"), upper case (``NOT'') have received the highest features weights ( Table \ref{table:topargfeatures}). In Figure \ref{figure:lstmcomp1}, while the individual model $LSTM_{attn}$ attends the words almost equally, we notice in the multitask variation   several sarcasm markers such as ``ya'', ``oops'', and numerous exclamations (``!!'')  receive larger attention weights.

Addressing the second issue (i.e., when both multitask and single tasks models make the wrong predictions), we notice that 
over 100 examples of none ($N$) class were classified as argumentative by both $BERT_{{MT}_{uncert}}$ and $BERT_{Orig}$. For the none $N$ class, one of the most common instances of wrong predictions is when the current turn $ct$ sarcastically takes a ``different stance'' on a topic from $pt$ in a narrow context but the whole turn is not argumentative. In the following example: ``does he just say the opposite of everything $<$name$>$ says?'' ($pt$) $\leftrightarrow$ ``using $<$name$>$ as a 180 compass is just fine by me'' ($ct$), $BERT_{{MT}_{uncert}}$,  $BERT_{Orig}$, LSTM$_{{MT}_{uncert}}$, and  $LSTM_{attn}$ models make disagree $D$ prediction (since $ct$ is sarcastic on ``$<$name$>$'')  where the gold label is none $N$. Looking closely at this pair of turns, it seems that the $ct$ presents a case of ad hominem attack (on the person's  ``$<$name$>$'') rather than a none relation. 

In the case of argumentative turns (agree and disagree) that are wrongly classified as none by all models, we found two common patterns: the use of concessions (e.g., ``it's a consideration, \emph{but} I doubt we should be promoting this \dots'') and arguments with uncommitted beliefs (e.g., ``it is \emph{possible} that'', ``that could \emph{probably} be'', ``\emph{possibly}, I must admit'').



\section{Conclusion and Future Work} \label{section:conclusion}

Linguistic and argumentation theories have studied the use of sarcasm in argumentation, including its effectiveness as a persuasive device or as a means to express an ad hominem fallacy. We present a comprehensive experimental study for argumentative relation identification and classification using sarcasm detection as an additional task.
First, in discrete feature space, we show that sarcasm-related features, in addition to argument-related features, improve the accuracy of the argumentative relation identification/classification task by 3\%. Next, we show that multitask learning using both a dual LSTM framework and BERT helps improve performance compared to the corresponding single model by a statistically significant margin. In both cases, the dynamic weighting of task specific losses performs best. 
We provide a detailed qualitative analysis by investigating a large sample manually 
and show what characteristics of sarcasm are attended to, 
which might have guided the correct prediction on the identification of the argumentative relation/classification task. 
In the future, we aim to study this synergy further by looking at sarcasm as well as the persuasive strategies (e.g., ethos, pathos, logos), and argument fallacies (e.g., ad hominem attack that was also noticed by \newcite{habernal2018before}). 

\section*{Acknowledgements}
The authors thank the
anonymous reviewers and Tuhin Chakrabarty for helpful comments. 
\bibliography{anthology,argusarc2020}
\bibliographystyle{acl_natbib}
\newpage
\section{Appendix} \label{label:appendix}


\subsection{Parameter Tuning}
\label{tuning}

\paragraph{Logistic Regression (LR) experiment:}

A Logistic Regression model with $L_{2}$ penalty is employed where the class weights are proportional to the number of instances for $A$, $D$ and $N$ classes. The regularization strength $C$ is searched over a grid using the $dev$ data. Following values were tried for $c$: [.0001, .001, .01, .1, 1, 10, 100, 1000, 10000].

\paragraph{Dual LSTM and Multi-task Learning experiment:} 
  
For LSTM networks based experiments we searched the hyper parameters over the $dev$ set. Particularly we experimented with different mini-batch size (e.g., 8, 16, 32), dropout value (e.g., 0.3, 0.5, 0.7), number of epochs (e.g., 40, 50), hidden state of different sized-vectors (100, 300) and the  Adam optimizer (learning rate of 0.01). Embeddings were generated using FastText vectors (300 dimensions) \cite{joulin2016fasttext}. Any token occurring less than five times were replaced by a special UNK token where the UNK vector is created based on random samples from a normal (Gaussian) distribution between 0.0 and 0.17. After tuning we use the following hyper-parameters for the $test$ set: mini-batch size of 16, hidden state of size 300, number of epochs = 50, and dropout value of 0.5. Task-specific losses for the dynamic multitask version was learned during training.

\paragraph{BERT based models:}
We use the $dev$ partition for hyperparameter tuning such as different mini-batch size (e.g., 8, 16, 32, 48), number of epochs (3, 5, 6), learning rate of 3e-5) and optimized networks with the Adam optimizer. 
The training partitions were  fine-tuned for 5 epochs with batch size = 16. Each training epoch took between 08:46 $\sim$ 9 minutes over a K-80 GPU with 48GB vRAM.

\subsection{Results on the Sarcasm Detection Task} \label{subsection:sarcresult}
Although improving sarcasm detection is not the focus our paper, 
we observe that multi-task learning improves the performance on this task as well, when compared to the single task model. We present results for the deep learning models in Table \ref{table:allsarcresults}. 
The multi-task models (both for LSTM and BERT) 
outperform the corresponding single task models (by 6.9 F1 and 6.4 F1 for LSTM and BERT models, respectively).
We note that the results on this particular dataset are much lower than on other datasets used for sarcasm detection.  For example, the LSTM$_{Attn}$ which is the best model used by \newcite{ghosh2018sarcasm}
obtained only 52.9 F1 score on this dataset, while it obtained 70.34 F1 on Sarcasm V2 (derived also from IAC but using different annotation guidelines), 74.96 F1 on a Twitter dataset and 75.41 F1 on a Reddit dataset \cite{ghosh2018sarcasm}.  

\begin{table}[h]
\centering
\begin{tabular}{cccc}
\hline
Model  & Precision &  Recall & F1 \\
\hline
{LSTM$_{Attn}$}  &  52.9 &	52.8&	 52.9 \\
{LSTM$_{MT}$}   & 59.5 &	59.3&	\textbf{59.4}  \\
\hline
 
{BERT$_{orig}$} & 57.4 & 57.4 & 57.4\\
{BERT$_{MT}$}  & 61.8 & 61.7 & \textbf{61.8} \\
{BERT$_{{MT}_{uncert}}$}  & 64.1 & 63.5 & \textbf{64.0} \\

\hline

\end{tabular}
\caption{Evaluations of sarcasm detection on the $test$ set of $IAC_{orig}$.}
\centering
\label{table:allsarcresults}
\end{table}

\fi

\end{document}